# Analysis of Agent Expertise in Ms. Pac-Man using Value-of-Information-based Policies

Isaac J. Sledge, *Student Member, IEEE* and José C. Príncipe, *Life Fellow, IEEE*

**Abstract**—Conventional reinforcement learning methods for Markov decision processes rely on weakly-guided, stochastic searches to drive the learning process. It can therefore be difficult to predict what agent behaviors might emerge. In this paper, we consider an information-theoretic cost function for performing constrained stochastic searches that promote the formation of risk-averse to risk-favoring behaviors. This cost function is the value of information, which provides the optimal trade-off between the expected return of a policy and the policy's complexity; policy complexity is measured by number of bits and controlled by a single hyperparameter on the cost function. As the policy complexity is reduced, the agents will increasingly eschew risky actions. This reduces the potential for high accrued rewards. As the policy complexity increases, the agents will take actions, regardless of the risk, that can raise the long-term rewards. The obtainable reward depends on a single, tunable hyperparameter that regulates the degree of policy complexity.

We evaluate the performance of value-of-information-based policies on a stochastic version of Ms. Pac-Man. A major component of this paper is the demonstration that ranges of policy complexity values yield different game-play styles and explaining why this occurs. We also show that our reinforcement-learning search mechanism is more efficient than the others we utilize. This result implies that the value of information theory is appropriate for framing the exploitation-exploration trade-off in reinforcement learning.

## 1 Introduction

Learning to successfully play a game is the process of discovering the correct actions to achieve both primary and secondary goals. Players typically do this in a trial-and-error-based manner: by taking risks, in the form of trying novel actions in particular game states, they begin to connect when to take certain actions. Eventually, given a diverse range of scenarios, players can relate sequences of actions and game states with the ability to obtain a given score. They can also refine the action sequences once this occurs.

There are a variety of machine-learning paradigms for mimicking this process. One example is reinforcement learning [1]. Reinforcement learning can be applied to sequential decision-making problems, such as games, in which the outcome of an action might not be immediately apparent. Reinforcement learning methods do this by assigning credit to actions leading to an outcome, which provides feedback in the form of expected rewards obtained from performing those actions. The expected rewards can then be manipulated to infer a mapping between environmental states and actions.

A great amount of reinforcement learning research has been conducted since the field's inception [2]. Despite these efforts, there are some shortcomings with many conventional approaches. For instance, there are few direct ways to guide the selection of risk-prone, -neutral, or -averse behaviors, as the learning process is typically driven by stochastic search. This applies to exploration heuristics such as epsilon-greedy and soft-max selection. Engaging in certain types of risky behaviors, at a certain frequency, may be necessary for agents to uncover high-performing policies. Moreover, it can be troublesome to construct policies that switch between different strategies during an episode when relying on many existing exploration heuristics, as we will show. This can occur because instances where different strategies are viable may be brief and rarely happen. If the agent avoids taking seemingly non-optimal actions in such instances, then it may seldom deviate from its current policy to determine if a new strategy would be appropriate under the given circumstances.

Both of these issues can be illustrated with the arcade game Ms. Pac-Man [3]. Ms. Pac-Man is a predator-prey scenario set in a maze-like environment. The game requires the use of multiple strategies to maximize its score. Throughout a majority of the game, the agent, Ms. Pac-Man, is being chased by invulnerable entities, which are ghosts. The

Isaac J. Sledge is with the Department of Electrical and Computer Engineering, University of Florida, Gainesville, FL 32611, USA (email: isledge@cnel.ufl.edu). He is also with the Computational NeuroEngineering Laboratory (CNEL) at the University of Florida.

José C. Príncipe is the Don D. and Ruth S. Eckis Chair and Distinguished Professor with both the Department of Electrical and Computer Engineering and the Department of Biomedical Engineering, University of Florida, Gainesville, FL 32611, USA (email: principe@cnel.ufl.edu). He is the director of the Computational NeuroEngineering Laboratory (CNEL) at the University of Florida.

The work of the authors was funded via grant N00014-15-1-2103 from the US Office of Naval Research. The first author was additionally funded by a University of Florida Research Fellowship, a Robert C. Pittman Research Fellowship, and an ASEE Naval Research Enterprise Fellowship.

objective of the agent is to clear the environment of pellets while navigating around the ghosts. However, after activating certain power-ups, the ghosts become vulnerable for a brief period of time. The agent can consume these ghosts for a score boost.

The switch in ghost dynamics necessitates a change in the game-play strategy, since multiple distinct modes of behavior are required under different conditions [4, 5]. Despite the need for multi-modal behaviors, conventional reinforcement-learning approaches have focused on constructing monolithic policies. Such policies would implement the same agent behaviors regardless of the vulnerability of the ghosts. Although it is possible to represent multi-modal behavior with these policies, it can be difficult to learn such behavior. This is, in part, due to risk. For instance, throughout the learning process, an agent may have learned to avoid colliding with the ghosts. Without straying from this behavior, the agent will not learn that there are instances where it can safely chase the ghosts.

In this paper, we consider an information-theoretic learning [6] approach for performing constrained stochastic searches that promote a continuum of risk-averse to risk-favoring agent behaviors during reinforcement learning. This, in turn, leads to a principled exploration of the state-action space that aids in the promotion of multi-modal behaviors. Other schemes, such as epsilon-greedy and soft-max selection, also implement action exploration. The novelty of our approach is that we frame exploration as a tractable optimization problem and understand the source of optimality. As well, existing theory shows that our approach leads to optimal decision making under uncertainty; similar guarantees are not known for other exploration mechanisms.

Our approach is based upon Stratonovich's value-of-information criterion [7–9], which provides an optimal conversion between rewards and information. Here, the goal is to analyze the evolution of the agent expertise, for the arcade game Ms. Pac-Man, when using the value of information.

The value of information possesses a single hyperparameter that specifies the policy complexity and influences the agent's game-playing expertise. This parameter arises automatically from optimization of the value of information. The policy complexity dictates the exploration granularity of the policy search space. If the complexity is low, then the policy space will be searched coarsely. The agent will implement cautious action-selection behaviors. That is, the agent will often avoid choosing actions that deviate from the policy. This can lead to low to modest long-term rewards if the policy does not accurately capture the environment dynamics. The amount of computational effort needed to learn in this case is low, however. When the policy complexity is high, the policy space is finely searched. The agent is prone to taking risky actions in this case. In doing so, the policies typically produce action sequences that lead to high long-term rewards. Such high-complexity policies are also likely to promote the formation of multi-modal behaviors. In either case, optimal-reward policies are found, where the actual best rewards are pre-determined by the policy complexity. This comes at the expense of computation, as many learning episodes may be needed for this to occur.

We show that distinct policy types, for Ms. Pac-Man, materialize for three ranges of parameter values. The agent merely clears the level of pellets and avoids the ghosts in the lowest-complexity case. It does not implement multi-modal behaviors. As the policy complexity is increased, the agent exhibits improved game-play styles, which come about from engaging in risky behaviors during training. For moderately low to moderately high policy complexities, the policies compel the agent to clear the pellets in an expeditious manner. It also has the agents chase after fruit power-ups. In the latter case, the agent switches between avoiding and sometimes chasing the ghosts whenever they are invulnerable and vulnerable, respectively. High-complexity policies attempt to produce the best rewards. Agents will continuously seek out power pellets, so as to remain invulnerable while clearing the level and pursuing nearby ghosts. The agents will also lure one or more ghosts to power pellets.

The remainder of this paper is organized as follows. We begin, in section 2, with a survey of the research that has been performed using Pac-Man and Ms. Pac-Man. Section 3 outlines our methodology. We first provide an overview of reinforcement learning using Markov decision processes. We then introduce the value of information from a Markov-decision-processes perspective. Risk-based interpretations of this criterion are provided. We also discuss implementational aspects. Our reinforcement learning simulations of the Ms. Pac-Man game are given in section 4. We begin this section by covering our experimental protocols. We then transition to our simulation results and corresponding analyses. These analyses include the qualitative and quantitative improvement in the agent's behaviors during training process. Performance comparisons against other reinforcement learning exploration strategies are provided to demonstrate the potential of our approach. Lastly, we conclude in section 5 and list possible avenues of future research.

## 2 Literature Review

Many approaches developed to play either Pac-Man or Ms. Pac-Man rely on computational intelligence principles [10–17]. The work of Lucas [10] is one of the earliest examples. He used evolutionary heuristics to train neural networks to play Ms. Pac-Man. His approach relied on the calculation of features relevant for gameplay, such as the distances to notable items and ghosts. A great deal of user intervention is needed for feature selection, though. Wittkamp et al. [11] also evolved neural networks, which were applied to learn ghost movement strategies for Pac-Man.

Their approach was based on the neuro-evolution of augmenting topologies paradigm. Using a manually-coded agent, they uncovered ghost strategies that can outperform those from the original game.

Evolutionary-inspired optimization tactics have also appeared in other works. Gallagher and Ryan [12] developed a finite-state machine approach to encode simple movement strategies. The movement strategies were evolved according to a population-based incremental learning algorithm. In [16], Gallagher and Ledwich described an approach to developing agents that learn the game from minimal on-screen information. The agents were created by evolving neural network controllers using straightforward evolutionary algorithms. Their results showed that neuro-evolution is able to produce agents that could play the game reasonably well, despite not having knowledge of the game rules. As indicated by the results in [4, 5], neuro-evolution frameworks can lead to agents with much higher performance. There is a complicated relationship, however, between the various hyperparameters and the expected agent behaviors for these types of methods. This makes it difficult to a priori assess the expected performance.

Alhejali and Lucas [14] used genetic programming to construct a variety of Pac-Man agents. Their results demonstrated that the agent behaviors functioned well for the training environment and could generalize well to novel environments. Brandstetter and Ahmadi [15] also relied on genetic programming, which was used as a reactive control mechanism for a Ms. Pac-Man agent. These evolutionary-inspired methods provide the ability to implement important game-playing behaviors. Nevertheless, there is currently little understanding about the connection between hyperparameter values and expected behaviors. Such methods also often do not leverage all of the agent's experiences when stochastically updating the policy during the crossover and mutation procedures. A great many training episodes may therefore be required before the stochastic search process uncovers a reasonable policy.

There have been a variety of non-computational-intelligence-, heuristic-based approaches that have been developed [18–20]. For instance, Szita and Lorincz proposed simple rule-based policies for controlling a Ms. Pac-Man agent. Their rules were organized into distinct action modules and the decision about which direction to move is based on the priorities associated with each action module. Policies for choosing the best action modules were uncovered using cross-entropy optimization. However, there is substantial user involvement in initially specifying the rules. More recently, Wirth and Gallagher [19] outlined how to utilize influence map models when playing Ms. Pac-Man. Their models highlight desirable and undesirable locations for the agents to visit in the game. They do not provide a means of learning influence maps from the agent's experiences, though. Human intervention is hence needed to identify parameter values for good maps.

Many of the above approaches can specify reasonably good agent behaviors. For some of them, though, it may be difficult to update the agent's response for a single situation or even a set of situations without also changing its behaviors for other instances. Some of them also may not adequately use the agent-environment interactions. Both issues complicate the efficient construction of good action-selection policies.

A means of resolving such issues is through the use of reinforcement learning. This style of learning has been employed to create Pac-Man and Ms. Pac-Man agents [21–26]. Such approaches also have the appeal that the user is out of the learning loop. The agent learns solely from its interactions with the environment, which mimics the biological archetype.

In [21], Burrow and Lucas applied interpolation-based temporal-difference learning to the task of constructing Ms. Pac-Man agents. Their experiments demonstrated that this style of learning led to more reliable performance than neuro-evolution. They also showed that the chosen reward structure can have a dramatic effect on the agent behaviors. In [23], Griffiths et al. addressed the problem of feedback integration with reinforcement learning. They showed that their style of policy-shaping-based reinforcement learning can be highly competitive compared to conventional learning approaches. Vezhnevets et al., in [24], applied deep neural networks for learning multi-step action strategies in a reinforcement learning setting. They showed that their agents strongly outperformed those derived from feed-forward neural networks and long short-term memory networks.

Another popular means of constructing agents has been through the use of Monte-Carlo tree search [27–31]. Robles and Lucas [27] applied a simple tree search heuristic on a Ms. Pac-Man agent to evaluate the danger of any particular course of action. Ikehata and Ito [28] applied upper confidence bounds to Monte Carlo search trees, which considers potential moves and formulates an average reward of action action sequence. Like the work of Robles and Lucas, that of Ikehata and Ito evaluates dangerous routes for the agent to take, based upon its location in the environment and any nearby ghosts. More recently, Samothrakis et al. [29] and Pepels et al. [31] applied Monte Carlo tree search to create high-performing agents. Foderaro et al. [32, 33] relied on tree searches as well. They first decompose the environment into a series of convex cells, which outlined locations where the agent could travel. The set of decomposed cells is used to create a decision tree for selecting a path that minimizes the risk of colliding with the ghosts.

Our approach for developing Ms. Pac-Man agents in this paper is related to those given in [21–26], as we also rely on reinforcement learning. It is also related to Monte-Carlo tree search [27–29, 31–33], since it can be viewed as a special case of reinforcement learning.

There are key differences, though. First, the action-selection scheme that we propose weights available infor-mation in the policy versus pay-off, which implicitly controls the quantization of the state-action decision space. That is, states are partitioned in accordance with the value function and a single action is assigned to each partition group. The number of partitions is dictated by the amount of risk, which is controlled by a single hyperparameter in the information-theoretic cost function. Provided that the amount of risk is properly chosen, this quantization functionality has the benefit of optimizing the efficiency of the learning process. We show that high-performing policies therefore emerge early during training. In contrast, policies that rely on conventional exploration strategies, which are used in some of the aforementioned works, lag in performance. This is because they must search over the original action policy space without an appropriate metric for the goal. Such exploration strategies also often fail to implement multi-modal behaviors, even when training over a great many episodes.

Another distinction is that we are combining information, which is a measure of uncertainty, with the rewards. As indicated by our simulation results, this appears to be a good framework to understand the variables at play for intelligent behavior. Our approach should also outperform existing reinforcement-learning methods that do not account for uncertainty in decision making, such as [10, 12, 14–16].

The reason why were able to obtain good performance is because the value of information combines the rigorous theory of utility, in the sense of von Neumann and Morgenstern [34], with uncertainty, as quantified by entropy. That is, the criterion is defined as a conditional extremum of expected utility under a relative entropy constraint. It therefore satisfies all of the axioms for expected utility and hence describes means of decision making under uncertainty that is independent of any specific definition of information [35]. It also does not, as noted by Belavkin [36], suffer from some of the perceived flaws of utility theory [37, 38]. This is because it includes a non-linear component that reflects the agent's preferences about potential information. Agents will therefore reason about expected improvements in the score differently than potential losses [39], which is necessary or optimal decision making.

Further advantages of our approach are provided in an associated online supplement. We also encourage readers to consult [40, 41] for more comparisons of the value of information with existing reinforcement learning strategies.

There are other minor advantages to our approach compared to some of the others that are used to provide Ms. Pac-Man agents. One example is that we have fewer parameter values that need to be chosen and tuned by the investigator. Two of the hyperparameters relate to the reinforcement learning updates, which are easy to specify based on previous studies. The remaining hyperparameter, which controls the amount of risk, is more difficult to set. Thus far, its influence appears to be application dependent. However, our simulation results provide insights into this hyperparameter, which will be useful in our future endeavors for understanding how it should be optimally chosen.

In our current work, we consider an annealing-based approach for setting the risk hyperparameter. This largely relieves the need for significant testing to a priori determine reasonably good values that are appropriate across all episodes. Provided that the annealing occurs at a sufficiently slow pace during training, existing reinforcement-learning theory ensures that the policy iterates will converge to the globally optimal action-selection strategy in the limit.

Convergence guarantees are another advantage of our approach. They indicate that the best-scoring agent will be uncovered after a sufficiently large number of state-action transitions. The more heuristic-based approaches like the ones in [18–20], in contrast, do not necessarily possess these guarantees. The same is largely true for Monte Carlo tree search [27–29, 31–33]. Convergence for many of these tree-search procedures is, currently, only guaranteed for simplified problem domains and for certain exploration strategies [30]. Their theoretical performance, in terms of a criterion known as regret, is not optimal in these domains, unlike the value of information [40].

## 3 Methodology
### 3.1 Markov Decision Processes

The value of information can be used for decision making when the agent-environment interactions are described by a general dynamical system. Here, we take this system to be a Markov decision process.

The Markov-decision-process framework is used heavily in the theory of stochastic decision-making for discrete-space and discrete-time settings [42]. In a Markov decision process, the system being considered is assumed to be driven by underlying Markov chains. That is, the system jumps randomly from one state $s_t \in \mathcal{S}$ to the next $s_{t+1} \in \mathcal{S}$ for discrete time steps $t$. The probability of transitioning from one state to another depends only on the current state and not on the history of states. Moreover, in a Markov decision process, the agent is required to choose an action $a_t \in \mathcal{A}$ from a set of available of actions. An immediate real-valued reward $g_{t+1} \in \mathcal{R}$ is earned during the transition between states $s_t$ and $s_{t+1}$ whenever an action $a_t$ is taken.

A corresponding policy dictates the action that is selected at each state. A policy $\pi_{s_{t+1}}(a_t, s_t)$ characterizes the probability distribution over actions given the state. We will sometimes use $\pi_{a_t}(s_t) : \mathcal{S} \to \mathcal{A}$ to denote the optimal action $a_t \in \mathcal{A}$ that is chosen for $s_t \in \mathcal{S}$ according to the current probabilistic policy. Each policy has an associated real-valued action-value function $q(s_t, a_t) : \mathcal{S} \times \mathcal{A} \to \mathcal{R}$. The goal of a reinforcement-learning agent is to find a policy that optimizes the value function for all state-action pairs: $\pi^*_{a_t}(s_t) = \arg\sup_{\pi_{a_t}(s_t):\mathcal{S}\to\mathcal{A}} \mathbb{E}[q(s, \pi_{a_t}(s_t))]$. Here, the value function is defined in terms of the discounted future rewards associated with a particular sequence of actions

$$q(s,a) = \mathbb{E}_{\pi_{a_t}(s_t)}\left(\sum_{t'=0}^{\infty} \gamma^t g_{t+t'+1}(s_{t+t'}, a_{t+t'}) \bigg| s_t = s, a_t = a \right), \quad (1)$$

for some discount factor $\gamma$ in the unit interval. The discounting factor ensures that the summation in (1) is finite.

The problem of finding the best policy can be written in terms of the reward function as

$$\sup_{\pi_{a_t}(s_t):\mathcal{S}\to\mathcal{A}} \left(\sum_{s_t\in\mathcal{S}} \sum_{a_t\in\mathcal{A}} p(s_t)\pi_{s_{t+1}}(a_t,s_t)q(s_t,\pi_{a_t}(s_t))\right) = \sum_{s_t\in\mathcal{S}} p(s_t) \sup_{a_t\in\mathcal{A}} \left(\pi_{s_{t+1}}(a_t,s_t)q(s_t,a_t)\right). \quad (2)$$

Markov decision processes are considered to be solved whenever the policy maximizes the future expected rewards from any starting state.

Markov decision processes are known to have an optimal policy. Finding this policy can be difficult, however. Most investigators therefore consider the construction of near-optimal policies, which can come about through the use of model-free, approximate dynamic programming approaches [43–45]. Model-free reinforcement learning approaches operate by constructing stochastic approximations of the value function in an attempt to circumvent the intractable nature of direct estimation. A mechanism behind this approximation is a selection of actions at each stage of the learning process. Actions can be selected based upon the current policy, which introduces no new information about the optimal value function. Novel actions, which deviate from the current policy, may also be chosen. This latter option leads to an exploration of the policy space and, oftentimes, the formation of accurate value-function approximations.

### 3.2 Value of Information

Our approach to the experience-based learning of policies for playing Ms. Pac-Man follows from the theory of Stratonovich [7, 8]. Stratonovich proposed a mathematical framework for bounded, rational decision making under information constraints. This took the form of an optimization criterion known as the value of information.

We show that the value of information can be used to determine when it is advantageous to deviate from the current policy. It therefore defines a type of reinforcement-learning exploration scheme, which, apart from setting a single hyperparameter, does not require user intervention. Our experiments will analyze the agent's expertise at playing Ms. Pac-Man when relying on different hyperparameter values.

The value of information is based on the premise that there is a trade-off between policy information and obtainable rewards. That is, the greater the amount of information, the better the long-term rewards should be. More time may be needed to find the optimal policy, though. Here, information takes the form of the number of bits to codify different policies. We sometimes refer to this as the policy complexity.

More specifically, we assume that an initial probabilistic action-selection policy is provided. The value of information is a constrained optimization problem that seeks a transformed version of this initial policy that yields the highest rewards. The initial policy is not necessarily allowed to change arbitrarily, though. Rather, the transformation is governed by a information-constraint term that assesses the divergence between the initial policy and the current policy. The total divergence cannot exceed a specified bound. The lower the bound, the less that the initial policy can be modified during learning. This will cause the agent to implement risk-adverse action-selection behaviors, which can impact the obtainable rewards if the initial policy does not encode the environment dynamics well. Conversely, the higher the information bound, the higher the diversity of policies and hence the greater the reward improvement. The agent will therefore favor risky actions. The degree of policy transformation must be delicately balanced to obtain the base possible rewards.

As we will show, the information bounds are dictated by a single parameter that emerges from converting this constrained optimization problem into an unconstrained optimization problem. This parameter simultaneously dictates how frequently actions that differ from the current policy are taken and hence how risky the agent's actions will be.

It is important to note that we are not the first to consider the application of information-theoretic concepts to reinforcement learning. For example, Tishby et al. [46, 47] proposed a Markov decision process formulation in which rewards are traded off against control information. In these papers, the authors noted that optimizing the amount of information yields Bellman-like equations, which they utilized to provide a dynamic programming formulation for solving the Markov decision processes. However, there is no attempt to couple information value, in the form of Shannon entropy, with rewards. Only a simple information bottleneck is implemented.

Despite that we also utilize information in our cost function, our approach differs in a few ways. For example, in the work of Tishby, they rely on control information, which corresponds to the amount of information gleaned by the agent about the environment from sensory perception. In our approach, information is related to policy transformation costs, which dictates the amount action-state-space exploration. Moreover, in [46, 47], the control information term becomes mixed with the action-value function, which complicates its estimation. Additionally, the value function no longer solely expresses discounted rewards, which makes its convergence analysis difficult. In our approach, we are still able to obtain an action-value function, and hence a policy, that reflects reward optimization. This allows us to easily apply existing convergence theory to value-of-information-based learning.

### 3.2.1 Quantifying Information Value

In the context of reinforcement learning, the value of information assumes that the agent's initial game-play behavior is defined by a prior distribution $\pi(a_t)$. Here, $a_t \in \mathcal{A}$ represents the action that that is chosen in the $t$th iteration; a single iteration, which we also refer to as an episode, is equivalent to taking a single action and transitioning from one state to the next.

By accruing experiences through interacting with the environment, the agent changes its game-play behavior to a posterior-like distribution $\pi_{s_{t+1}}(a_t, s_t), a_t \in \mathcal{A}$ and $s_t \in \mathcal{S}$, which is the probabilistic policy. This change occurs in a way that optimally trades off the expected utility against the transformation costs for going from $\pi(a_t)$ to $\pi_{s_{t+1}}(a_t, s_t)$. The trade-off is given by the following extremization problem, which is the value of information for the Shannon information case,

$$- \underbrace{\sup_{a_t \in \mathcal{A}} \left( \sum_{s_t \in \mathcal{S}} p(s_t) q(s_t, a_t) \right)}_{\sup_{a_t \in \mathcal{A}} \left( \mathbb{E}_{p(s_t)}[q(s_t, a_t)] \right)} + \underbrace{\sup_{\pi_{s_{t+1}}(a_t, s_t)} \left( \sum_{s_t \in \mathcal{S}} \sum_{a_t \in \mathcal{A}} p(s_t) \pi_{s_{t+1}}(a_t, s_t) q(s_t, \pi^*_{a'_t}(s_t)) \right)}_{\sup_{\pi_{s_{t+1}}(a_t, s_t)} \left( \mathbb{E}_{p(s_t)} [\mathbb{E}_{\pi_{s_{t+1}}(a_t, s_t)}[\sup_{a_t \in \mathcal{A}} q(s_t, a_t)]] \right)}; \quad (3)$$

here, we use $\pi^*_{a'_t}(s_t)$ to denote the optimal action $a'_t \in \mathcal{A}$ that is taken in state $s_t \in \mathcal{S}$ according to the optimal policy. The optimization of the probabilistic policy is subject to the constraint

$$\pi_{s_{t+1}}(a_t, s_t) \text{ such that}: \underbrace{\sum_{s_t \in \mathcal{S}} p(s_t) \sum_{a_t \in \mathcal{A}} \pi_{s_{t+1}}(a_t, s_t) \log \left( \frac{\pi_{s_{t+1}}(a_t, s_t)}{\pi(a_t)} \right)}_{D_{\text{KL}}(\pi_{s_{t+1}}(a_t, s_t) \| \pi(a_t))} \leq \varphi_{\inf}, \; \varphi_{\inf} > 0, \quad (4)$$

that measures the amount of information that the prior contains about the policy. This expressions in (3) and (4) can be viewed as problem of finding a version $\pi_{s_{t+1}}(a_t, s_t)$ of the prior $\pi(a_t)$ that achieves maximal rewards in the case where $\pi(a_t)$ can only diverge by a prescribed amount $\varphi_{\inf}$.

The constrained criterion given by (3) and (4) can be converted into an unconstrained variational expression. This unconstrained maximization problem follows from the theory of Lagrange multipliers. The expression is composed of the expected utility of the agent under the posterior policy minus the information transformation cost that is required for forming the posterior,

$$\underbrace{\left( \sum_{s_t \in \mathcal{S}} \sum_{a_t \in \mathcal{A}} p(s_t) \pi_{s_{t+1}}(a_t, s_t) q(s_t, \pi^*_{a'_t}(s_t)) \right) - \frac{1}{\vartheta_t} \sum_{s_t \in \mathcal{S}} p(s_t) \sum_{a_t \in \mathcal{A}} \pi_{s_{t+1}}(a_t, s_t) \log \left( \frac{\pi_{s_{t+1}}(a_t, s_t)}{\pi(a_t)} \right)}_{\vartheta_t^{-1} \log \left( \sum_{a_t \in \mathcal{A}} \pi(a_t) e^{\vartheta_t q(s_t, a_t)} \right)}, \; \vartheta_t > 0, \quad (5)$$

for some positive $\vartheta_t$ that is allowed to change across iterations. The parameter $\vartheta_t$ sets the relative importance between the transformation cost and the reward maximization. Adjusting $\vartheta_t$ across each iteration modifies the amount of policy-space exploration that is performed and hence the riskiness of the agent's action-selection behaviors.

**Criterion Explanation.** The two terms in (3) define a difference in expected rewards. They can be described as follows:

    **First Term: No-Information Returns.** The first term captures the possible returns $\sup_{a_t \in \mathcal{A}} \mathbb{E}_{p(s_t)}[q(s_t, a_t)]$ for a policy in which the divergence between the states and actions are zero. That is, the states carry no information about what action should be taken. This is used to establish the baseline agent performance.

    If the states are not informative, then the optimal action is based solely on the state random-variable distribution and the action-value function. If, however, the actions are informative, then the returns for the simplest policy will be offset by a second term.

    **Second Term: Informative Returns.** This second term quantifies the returns associated with policies whose transformation costs are non-zero. It is based on the expected return using a modified action-value function, $\sup_{\pi_{s_{t+1}}(a_t, s_t)} \mathbb{E}_{p(s_t)} [\mathbb{E}_{\pi_{s_{t+1}}(a_t, s_t)}[\sup_{a_t \in \mathcal{A}} q(s_t, a_t)]]$. The optimal action-selection strategy $\pi_{s_{t+1}}(a_t, s_t)$ is initially unknown. However, it is assumed to be related to the specified prior $\pi(a_t)$, in some manner, as described by the information-constraint bound.

For this term, the divergence between states and actions is assumed to be non-zero. States therefore carry some information about what action should be chosen, which allows for the formation of a non-uniform policy.

As the number of training episodes becomes unbounded, the agent will have complete knowledge of the environment, assuming that the information-constraint bound $D_{\text{KL}}(\pi_{s_{t+1}}(a_t, s_t) \| \pi(a_t)) \leq \varphi_{\text{inf}}$ is equal to the state random variable entropy; that is, $\varphi_{\text{inf}} = -\sum_{s_t \in \mathcal{S}} p(s_t) \log(p(s_t))$. This second term will eventually produce returns that converge to those from the full-information case. Optimal pay-outs can also be achieved if the information constraint bound is annealed at a sufficiently rapid pace. The agent's action-selection behavior becomes entirely deterministic in such situations, as only the best-performing action will be chosen. The optimal policy becomes a delta function in the limit. Otherwise, the agent's behavior will be semi-random. That is, the policy will be described by a discrete distribution that is not quite uniform and not quite a delta function.

The constraint term in (4) can be described as follows:

**Constraint Term: Transformation Cost.** The constraint term, $D_{\text{KL}}(\pi_{s_{t+1}}(a_t, s_t) \| \pi(a_t))$, a Kullback-Leibler divergence, quantifies the amount of informational overlap between the posterior $\pi_{s_{t+1}}(a_t, s_t)$ and the prior $\pi(a_t)$. This term is bounded above by some non-negative value $\varphi_{\text{inf}}$, which implies that the amount of overlap will be artificially limited by the chosen $\varphi_{\text{inf}}$.

The value of $\varphi_{\text{inf}}$ dictates by how much the prior $\pi(a_t)$ can change to become the posterior $\pi_{s_{t+1}}(a_t, s_t)$. If $\varphi_{\text{inf}}$ is zero, then the transformation costs are infinite. The posterior will therefore be equivalent to the prior and no exploration will be performed; the agent's action-selection behaviors will be extremely cautious. The expected reward will not be maximized for non-optimal priors. If $\varphi_{\text{inf}}$ is larger than the state random-variable entropy, then the transformation costs are ignored. The prior is free to change to the optimal-reward policy, since the exploration of actions dominates. The agent's action-selection is therefore incredibly risky and hence may differ from the current policy. For values of $\varphi_{\text{inf}}$ between these two extremes, the agent weighs the expected improvement in rewards against the transformation costs. A mixture of new-action exploration and exploitation of the current best-scoring actions will take place. Some amount of risk-taking behaviors will emerge.

The constraint can also be viewed as describing the amount of information available to the agent about the environment. That is, for small values of $\varphi_{\text{inf}}$, the agent has little knowledge about the problem domain. It is unable to determine how to best adapt its action-selection strategy. As $\varphi_{\text{inf}}$ increases, more information is made available to the agent, which allows it to refine its behavior. In the limit, the agent has the potential for complete knowledge about the environment, which means the action choice will be reward-optimal.

The value of information, as defined by either (3) and (4) or (5), therefore describes the best pay-out that can be achieved for a specified policy transformation cost.

**Criterion Risk Promotion.** The expression in (3) defines an upper frontier on utility. We can write this upper frontier in the form shown in (5): $\kappa_{\vartheta_t}(\pi_{a_t}) = \vartheta_t^{-1} \log(\sum_{a_t \in \mathcal{A}} \pi(a_t) e^{\vartheta_t q(s_t, a_t)})$. We also define a lower frontier on utility, which corresponds to the worst-case scenario of minimal reward. In either case, the optimizations are subject to an information bound (4). These two frontiers stem directly from the theory and are referred to as the abnormal and normal branches. They represent the maximal gain and maximal reduction in reward over the prior, respectively.

Properties of these two branches give insights into the riskiness of the action-selection behaviors. Toward this end, we note that action-state spaces with non-zero entropy have a non-zero information potential. This means that after taking an action, the Shannon information may increase or decrease by the some amount, which corresponds to changing the information bound by some amount:

**Low Information Bound.** If the constraint is decreased, then the value of information will shift toward the normal branch. The normal branch implies risk-aversion in action choices [36]. This is because the potential increase of $\kappa_{\vartheta_t + \vartheta_t'}(\pi_{a_t}) - \kappa_{\vartheta_t}(\pi_{a_t})$, associated with an increase in $\vartheta_t$, is less than the potential decrease of $\kappa_{\vartheta_t}(\pi_{a_t}) - \kappa_{\vartheta_t - \vartheta_t'}(\pi_{a_t})$, associated with a decrease in $\vartheta_t$. Empirical and theoretical evidence indicates that risk-aversion is related to a quantization of the state-action space in homogenous risk regions. That is, as the information constraint parameter is decreased, many states will be grouped together. Visiting a particular state group will elicit the same action response. Sub-groups of states may be uncoupled and assigned a different action response during the learning process, but this rarely occurs. Consequently, there is little impetus for an agent to undertake risky actions, as once a policy has been found that leads to a reasonably high score, it is usually not modified much.

**High Information Bound.** If the information constraint parameter value is increased, then the value of information will shift away from the normal branch and toward the abnormal branch. The convexity of the abnormal branch promotes risk-taking behaviors when choosing actions [36]. This is because the potential increase $\kappa_{\vartheta_t}(\pi_{a_t}) - \kappa_{\vartheta_t + \vartheta_t'}(\pi_{a_t})$ is greater than the potential decrease $\kappa_{\vartheta_t - \vartheta_t'}(\pi_{a_t}) - \kappa_{\vartheta_t}(\pi_{a_t})$. Our experiences indicate that the action-state space is finely quantized when this occurs; ranges of the information constraint parameter give rise to different policies. Every state therefore has the potential to be paired with a unique action. Groups of states that are formed early in the learning process may be uncoupled in the later stages. Taking actions that deviate from the current policy is naturally more probable, as they will alter the state-action mapping and lead to potentially higher-reward policies.

As noted by Belavkin [36], risk is inherently asymmetric: rational agents should act differently when taking risks versus when they are avoiding risks. The value of information naturally implements this behavior when weighting action choices, as it produces an asymmetric value function mirroring the one found in prospect theory [39].

### 3.2.2 Finding Policies using Information Value

The preceding form of the value of information criterion has some practical difficulties. In particular, investigators must have knowledge of the maximum expected reward associated with the globally optimal policy. Even just estimating this amount can be troublesome for many environments. We demonstrated, in [48], that the criterion could be reformulated, in an equivalent manner, to sidestep this issue.

As a byproduct of this reformulation, we obtain a constrained convex criterion that can be converted into an unconstrained one using the theory of Lagrange multipliers. This unconstrained criterion can be globally optimized according to an expectation-maximization-like update, which define a weighted-random exploration scheme for reinforcement learning.

This exploration strategy can be combined with a model-free reinforcement learning approaches. Here, we have use tabular $Q$-learning, which relies on the estimation of action-value functions for each state-action combination. The resulting discrete, probabilistic policy can be found from these value functions, in an iterative manner, as shown in algorithm 1. We assume that all of the actions for a given state in the policy are initialized to have discrete uniform probabilities.

The corresponding optimization steps for the value of information are given in algorithm 1, steps 7 and 8. Step 7 is the expected value of the conditional action-selection probabilities over all states. It defines the average action-selection probability. The update in step 8 leads to a process for weighting action choices in each state according to a Boltzmann distribution parameterized by the expected reward. This resembles soft-max-based action selection. The difference is that an extra term, whose form is given in step 7, has been included to account for the promotion or suppression of risky actions. This term, along with an associated parameter, weights the expected rewards according to the level of risk that an investigator wishes an agent to possess. It is straightforward to show that this action weighting is greedy in the limit, under some relatively mild conditions. Therefore, it will produce optimal-reward policies [49].

---

**Algorithm 1:** Value-of-Information-Based $Q$-Learning

1. Choose unit-interval values for the discount and step-size parameters $\gamma_t \in \mathcal{R}_{0,1}, \alpha_t \in \mathcal{R}_{0,1}$.
2. Choose non-negative values for the agent risk-taking parameters $\vartheta_t$.
3. Initialize the action-value function $q(a,s)$ and hence the policy $\pi(a,s)$, $s \in \mathcal{S}$, $a \in \mathcal{A}$.
4. **for** $t = 0, 1, 2, \ldots$ **do**
5.     For all relevant $s_t \in \mathcal{S}$ and $a_t \in \mathcal{A}$, update the state probabilities
$$p(s_t) \leftarrow p(s_0) \sum_{s_0 a_0, \ldots, s_{t-1} a_{t-1}} \prod_{j=1}^{t} p(s_j | a_{j-1}, s_{j-1}) \pi_{s_j}(a_{j-1} | s_{j-1}).$$
6.     **for** $k = 0, 1, 2, \ldots$ **do**
7.         Update $\pi^{(k)}(a_t) \leftarrow \sum_{s_t \in \mathcal{S}} \pi_{s_{t+1}}^{(k)}(a_t, s_t) p(s_t)$, $a_t \in \mathcal{A}$.
8.         For $s_t \in \mathcal{S}$, $a_t \in \mathcal{A}$, update
$$\pi_{s_{t+1}}^{(k+1)}(a_t, s_t) \leftarrow \pi^{(k)}(a_t) e^{q(s_t, a_t)/\vartheta_t} \bigg/ \sum_{a_t \in \mathcal{A}} \pi^{(k)}(a_t) e^{q(s_t, a_t)/\vartheta_t}.$$
9.     Choose an action, in a weighted random fashion, according to the policy $\pi_{s_{t+1}}^{(k+1)}(a_t, s_t)$; perform a state transition $s_t \to s_{t+1} \in \mathcal{S}$. Obtain a reward $g_{t+1} \in \mathcal{R}$.
10.     Update the action-value function estimates $q(a_t, s_t) \leftarrow$
$$q(s_t, a_t) + \alpha_t \left( g_{t+1}(s_t, a_t) + \gamma_t \sup_a q(s_{t+1}, a) - q(s_t, a_t) \right).$$

---

The inner loop of algorithm 1, given from steps 7 to 8, covers the expectation-maximization-like update of the policy according to the value of information. We have shown that the policy approximation error for this loop is proportional to one over the number of updates; here, we consider only ten updates. The outer loop of algorithm 1, given from steps 5 to 10, defines the main reinforcement-learning process. In our experiments, we iterate over this loop for only a few thousand episodes. Note that alternate stopping criteria for both of these loops can be defined based upon reaching some steady-state solution.

The parameter $\vartheta_t$ that arises upon minimization of the unconstrained value of information dictates the rate of change the reward difference with respect to the Shannon information constraint. As the value of the parameter goes to infinity, the policy complexity constraints are ignored. As the value of the parameter goes to zero, the relative entropy between the actions and states is minimized. This implies that all states elicit the same response, since the transformation costs are so high that the agent cannot change its behavior. Despite this limitation, though, the agent will still select actions that attempt to optimize the expected policy returns. For values of the rationality parameter between these two extremes, there is a trade-off that occurs. As the parameter value increases, the agent may favor state-specific actions that yield a better expected return for particular states. As the parameter value decreases, the agent will favor actions that yield high rewards for many observations, which produces a lower relative entropy.

Choosing good parameter values can be challenging, as their influence over agent risk is application dependent. There are a few strategies that can be used for this purpose, though. The option that we espouse is to apply an annealing proces to iteratively adjust the amount of agent risk. Under this scheme, an initial value for $\vartheta_0$ could be selected for the parameter. This value can be chosen using expert knowledge so that there is potential for reward improvement. After each episode, the value of $\vartheta_t$ would then be decreased. This forces the agent to take less risks as the learning progresses. This adjustment process repeats until the parameter reaches some minimal value. By this time, assuming that the annealing rate was not too quick, the agent should choose high-reward actions in a deterministic manner.

## 4 Simulations and Discussions

We now assess our reinforcement learning strategy for a Ms. Pac-Man agent, which is a sufficiently challenging problem. Foremost, it is a dynamic environment in which there are moving obstacles, which are the ghosts. The ghosts' movements are random, which makes the entire problem stochastic. As well, the game requires switching between different play styles, upon consuming power pellets, so as to maximize the total rewards. This shift in play styles is predominantly risk driven. Without an ability to promote risky agent behaviors, the development of multi-modal strategies may not occur.

The aims of our reinforcement learning simulations are multi-fold. First, we would like to quantify the influence of the information constraint parameter on the agent's game-playing expertise. Secondly, we want to determine how much risk is needed for multi-modal behaviors to emerge. Both of these aims help in understanding the role of the risk hyperparameter. Lastly, we want to see how the behaviors obtained through use of the value of information compare with those from other reinforcement-learning exploration strategies.

### 4.1 Simulation Preliminaries
#### 4.1.1 Agent Rewards

We utilize a dual scoring system for our simulations. The first score that we report is directly from the Ms Pac-Man game environment. We refer to this as the original score in our plots. For this score, the agent earns +10 and +50 points for consuming a pill and power pill, respectively. The values of bonus items range from +100 to +5000 points, depending on the level. The points obtained for consuming ghosts increases exponentially, +200, +400, +800, +1600, according to the number of ghosts eaten during the effects of a single power pellet.

For the purposes of reinforcement learning, we rely on a modification of the above scoring scheme. We refer to this as the modified score in our plots. In particular, a negative reward is incurred whenever Ms. Pac-Man moves in one of four cardinal directions to a new location ($-1$). Remaining within the same location across a state transition is also penalized ($-1$). Both of these penalties spur the agent to avoid unnecessary backtracking and wandering. Colliding with a ghost, without having recently eaten a power pellet, results in a high penalty ($-500$). This is done to ensure that the agent implements ghost avoidance behaviors early during training. All of the other rewards are the same, except those for eating vulnerable ghosts. We quadruple the points to better promote the formation of multi-modal strategies.

#### 4.1.2 State Spaces

There are a variety of features that can be used to codify the decision-process state space. Here, we compute a series of holistic features. Such features include the distance to the nearest pellet, power pellet, fruit, nearest ghost, and second nearest ghost, all of which are in relation to the agent. These distances are measured by the number of steps the agent would need to reach those locations in the most direct fashion. We quantize the distances into two categories: near and far. We also include the cardinal direction in which the agent would need to travel to reach those locations, assuming no obstacles are present in the level. Directional ties are broken randomly. We also have binary features that determine if the two nearest ghosts are in the same tunnel as the agent, if the fruit has spawned, if a power pellet is active, if the nearest and second nearest ghosts are vulnerable, and if the agent is in a tunnel or a junction. Lastly, we have a feature that quantifies the remaining steps before the power pellet wears off. In total, there are eighteen features.

We chose these holistic features, as they encode important details about the game-play mechanics. They can also be quantized significantly without much loss in the agent's decision-making capability. Such global features therefore facilitate the formation of good-performing policies in relatively few training episodes. Comparatively, features that rely on more local information tend to require significantly more training episodes to reach the same performance levels. As well, the holistic features are not dependent on the level configuration. Policies obtained from them should hence generalize well to a variety of situations.

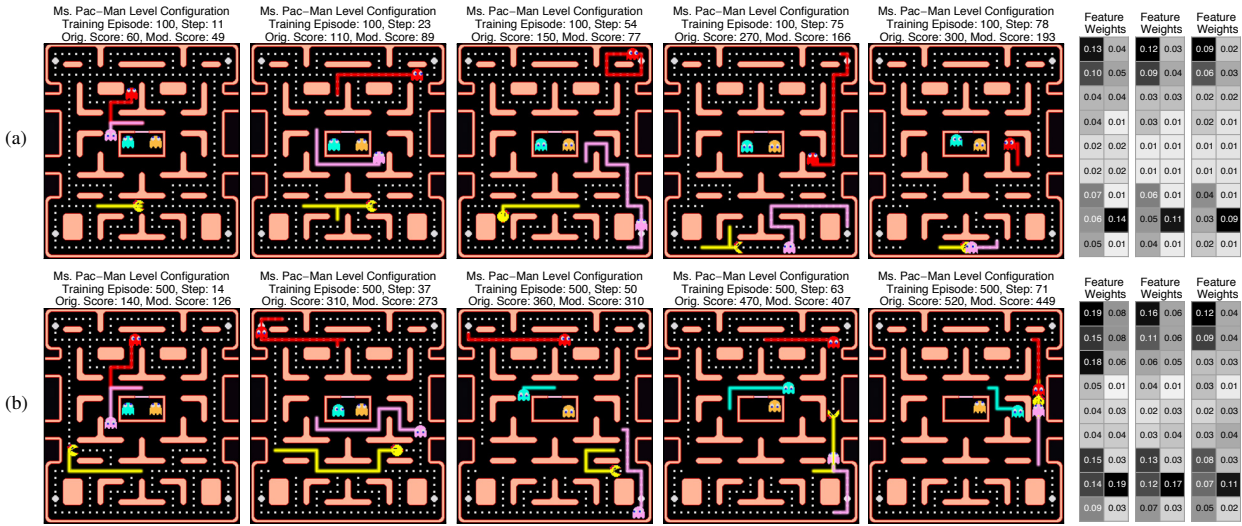

Figure 1: An overview of the agent's ghost avoidance behavior improvements over multiple training episodes. In row (a), the agent does not yet possess knowledge of the dynamics of the environment. It therefore moves about seemingly at random and often collides with ghosts after only a few steps. In row (b), the agent has learned to move around the level less haphazardly so as to consume more pellets. It still sometimes collides with ghosts, however.

The images in rows (a) and (b) on the left-hand side of the figure highlight the environment state over multiple actions for a given training episode. These images were obtained from the moderate-complexity case. In each of these images, the colored lines indicate the paths that the agent (yellow line) and the ghosts (red, pink, cyan, and orange lines) took through the environment across a number of action choices. At the end of each row, we provide a table of the normalized feature weights for each of the eighteen holistic features. These tables are the features for the low-, moderate-, and high-complexity policies, respectively. Darker colors indicate greater feature importance. The features are presented in the same order as they were introduced from the discussions, where the numbering is from top to bottom then left to right. A policy complexity of 3760 bits was used for the results given in rows (a) and (b).

### 4.2 Simulation Results

The simulation results that we present are obtained from Microsoft's Revenge of the Arcade software package, which is identical to the original arcade version of Ms. Pac-Man. We use the color- and pattern-based screen capture methodology described in [32] to interact with this application.

We assume that each episode provides the agent with five lives, instead of the usual three. A new life can be gained once per episode in accordance with the game's natural scoring system. A life is lost if the agent collides with an invulnerable ghost. A training episode ends if the agent has no lives.

We have three free parameters that need to be set during the learning process. The first parameter determines the amount of policy-space search granularity. Based upon initial simulations, we have found that a reasonable value for this is 3760 bits. This value is the halfway point between a minimal policy complexity, 1880 bits, and the maximal policy complexity that we consider, 5640 bits. The minimal policy complexity describes the least number of control bits needed to clear the given level, assuming a minimal amount of backtracking is performed. The maximal policy complexity was obtained assuming that the agent may have to visit each location in the entire level three times. A parameter value of 3760 bits should therefore allow the agent to evade ghosts and still complete the level. It should also strike a good risk balance.

For all of our experiments, we anneal the policy complexity from its initial value toward zero. We do this by reducing the policy complexity by anywhere from 0.01 to 0.05% of its current value across each episode.

The remaining two parameters are the learning rate and the discount factor, which dictate the action-value update behavior when using $Q$-learning. We set the learning rate to 0.3 so that more weight is given to previously acquired action-value magnitudes than those that were more recently acquired. We temporarily increase the learning rate to 0.7 for a single episode if that episode has the highest reward compared to all previous episodes. This ensures that the policy quickly implements new agent behaviors to better play the game. A discount factor to 0.9 was used so that the agent would seek action sequences with high long-term rewards. Each of these values was chosen based upon previous applications of $Q$-learning to stochastic game environments.

#### 4.2.1 Emergent Agent Behaviors

The first aspect of the value of information that we consider is the relationship between risk, as dictated by the policy complexity, and the emergent agent behaviors. That is, we look to see what strategies are implemented by the policies when the number of bits used to represent them changes. We also assess how these behaviors are altered for different risk amounts.

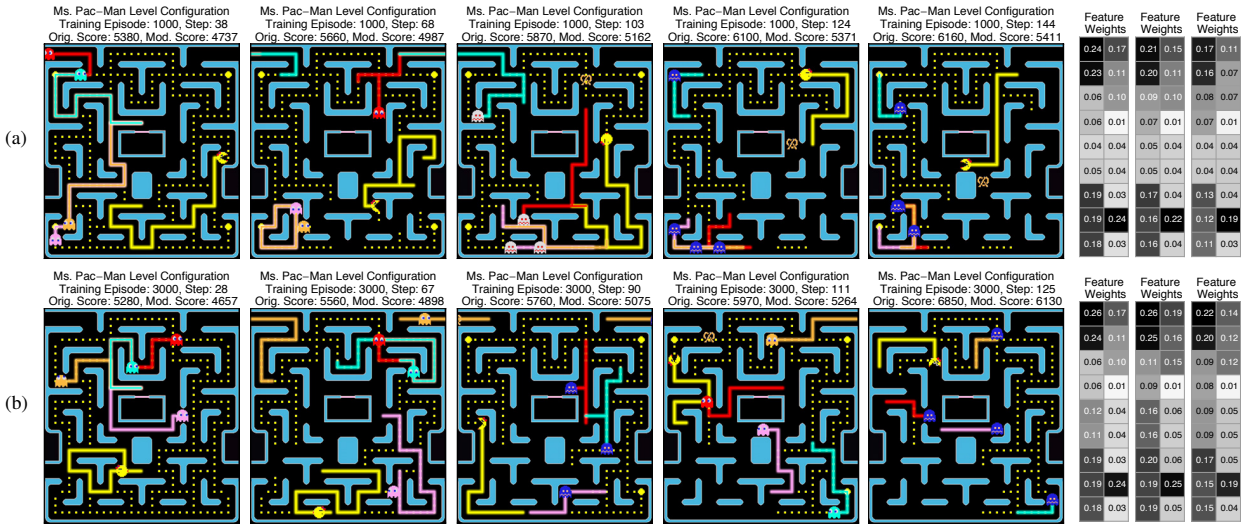

(a)

(b)

Figure 2: An overview of the agent's fruit chasing improvements over multiple training episodes. Before a few thousand episodes, the agent often does not know that fruits are a source of extra points. It therefore ignores them, even when they are only one or two steps away. This is highlighted in row (b). Through random chance, the agent will eventually consume fruits. If this occurs frequently enough, the agent will update its policy to implement fruit-chasing behaviors. As shown in row (b), the policy will direct the agent to consume fruits even when they are somewhat far away. We refer readers to the caption in the first figure for information about each of the plots.

**Basic Game-Play Behaviors.** We encountered three sets of behaviors at different stages of the learning process, depending on the policy complexity. The first set of behaviors focused on the tasks of avoiding ghosts and seeking pellets. This functionality emerged over 500 to 1000 episodes and was refined across another 1000 to 3000 episodes, depending on the policy complexity.

Figures 1(a)–1(b) give a visual overview of the agent's behavior at different points during the first 500 episodes. As shown in figure 1(a), the agent's initial performance is poor, regardless of the policy complexity. The agent often either remains stationary or executes unnecessary actions, increasing the agent's chance to be eaten by ghosts. Additionally, the agent has not yet learned that colliding with ghosts results in a large, negative response. The agent therefore clears only small sections of the level. Such behaviors arise since many relevant value-function entries do not yet have meaningful values.

As the agent accrues more experiences, the action-value function estimates become more accurate. The agent can thus more effectively clear the level. This is highlighted in figure 1(b). The results in figure 1(b) demonstrate that the agent spends less time uselessly backtracking through already cleared sections of the level. The agent also actively avoids ghosts. However, the agent tends to not keep a conservative distance between itself and nearby ghosts, as indicated in tables A.1(a)–A.1(c) of an associated online supplement. This can lead to it being trapped when the ghosts launch a two- or three-pronged pincer attack, as is the case in figure 1(b). As shown in tables A.1(a)–A.1(c), later in the learning process, the agent increases the average distance between it and the ghosts. Many flanking strategies that the ghosts employ are no longer effective at this point, since the agent typically leaves enough of a gap to escape.

These behaviors emerged due to an increased utilization of the state-space features. We have included plots of the normalized feature weights as found using the approach of Emigh et al. [50]. The weights indicate the importance of the features for decision-making, with higher values implying greater importance. The features are listed from top to bottom, left to right in the order that they were introduced in the previous subsection.

As shown in figure 1(a), the weights are mostly homogeneous for the three policy complexity cases by episode 100. This suggests that there is little preference, with some exceptions, for using various features to take certain actions. By episode 500, more emphasis is given to features related to the distance and position of the nearest pellet and nearest ghosts, if the agent is in a tunnel, and if the ghosts are in the same tunnel as the agent. This is highlighted in figure 1(b). Such changes correspond with the ability for the agent to clear larger portions of the level. Note that lower-complexity policies have larger feature weights than policies of higher complexity by this time, implying that they better account for the environment dynamics. After around 1500 to 3000 episodes, policies of higher complexity will have larger weights and hence yield better scores.

The rate at which these basic game-playing skills are acquired is based on the policy complexity. Low complexities, which for this application are around 1500-2500 bits, often implement ghost-chasing and pellet-seeking behaviors more quickly than policies with a higher complexity. This is due to an inherent quantization of the action-state space. Many states are grouped together, in this case, and actions can be chosen that work reasonably well for all of the states in a given group. Only a moderate amount of agent experience over a few hundred to thousand episodes is needed to determine which actions are appropriate. Higher-complexity policies, which range from 2600-6600 bits

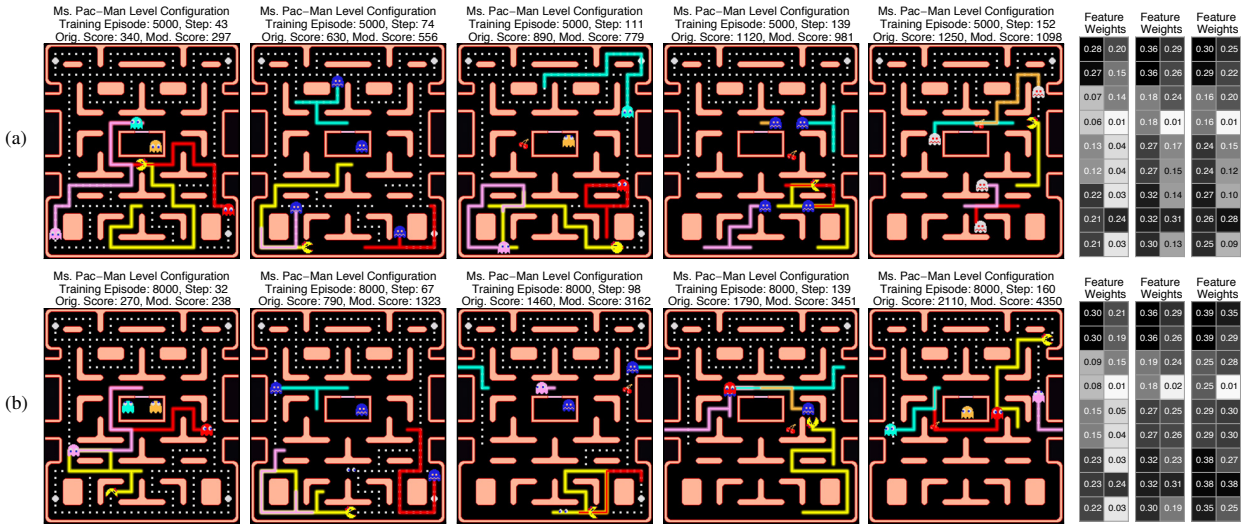

Figure 3: An overview of the agent's ghost-chasing behavior improvements over multiple training episodes. Initially, the agent does not know that vulnerable ghosts can be eaten. It therefore avoids them, as shown in (a). After several thousand episodes, the agent will sometimes consume vulnerable ghosts. This can happen, for instance, when the agent is stuck between two or more ghosts with no path for escape. If the agent consumes enough ghosts, the policy will begin to associate such actions with increased rewards. It will then implement multi-modal behaviors that depend on the ghost state. As shown in (b), the agent will eventually chase after vulnerable ghosts.

We refer readers to the caption in the first figure for information about each of the plots.

for this application, require many more agent-environment interactions to reach the same skill level. This is because the state-action space quantization is much finer, which leads to more state groups with fewer states. The agent is also more risky with its action-selection strategy, so poor choices may be made early in the learning process. However, as we highlight in the next subsection, this greater amount of risk offers the opportunity to refine these behaviors.

**Fruit-Chasing Behavior.** The next set of behaviors focused on chasing spawned fruit. Consuming fruits provide varying amounts of points: it starts from a reward of $+100$ and increases to $+5000$ as more levels are cleared before the agent loses all of its lives.

During the first 2000-3000 episodes, the agent often ignores any spawned fruit that maneuver through the level, regardless of the policy complexity. This is because the policy has not yet associated eating fruit with higher scores. In fact, the agent will not pursue fruit, even if it is adjacent to the agent. This is highlighted in figure 2(a).

Eventually, by coincidence, the agent will be in the same corridors as fruits and consume them. By about episodes 3000 to 4500, this will have happened frequently enough for the agent to actively chase fruits from a few cells away, assuming that there are no ghosts along that path. Supporting statistics are provided in tables A.2(a)–A.2(c). This reliably occurs only for moderate- and high-complexity policies, though. The transformation costs for low-complexity policies are too great to consistently implement this behavior; the scores for such policies tend to lag by this point as a result, which we show in the next subsection.

After anywhere from 6000 to 9000 episodes, the agent will seek out fruits at a greater distance. It will even do so if the chosen path has few pellets along it, since the score improvements offset the movement penalties. Supporting statistics are given in tables A.2(a)–A.2(c), which show that high-complexity policies tend to chase fruits over longer distances than moderate-complexity policies. Beyond a certain distance, the agent tends to overlook such bonus items, especially if it is within the same corridor as a ghost. The feature weights in figures 2(a) and 2(b) signal that this happens because ghost avoidance is a more prominent behavior. Eventually, knowing the location of both fruit and ghosts become equally important for the purposes of maximizing the rewards.

The agent has three features it can use to determine when a fruit has spawned. The first two are related to the distance and direction of the fruit in the environment, which has special values for non-existent fruit. The other is a binary-valued feature that determines if a fruit has spawned or not. As shown in figures 1 to 3, this latter feature consistently has a low weight, even after many episodes. It is therefore not important for decision making, as similar information can be inferred from the other features. We found that the fruit-spawned binary feature could be removed without adversely impacting the agent's performance. In many cases, when re-starting the learning process using this reduced feature set, a slight increase in the rate of skill acquisition was observed for the three different policy complexities.

**Ghost-Chasing Behavior.** Another behavior that materializes is that of eating vulnerable ghosts. This adds anywhere from $+200$ to $+1600$ points to the original score and $+800$ to $+6400$ points to the modified score, depending on the number of ghosts consumed per power pellet.

Initially, the agent does not know if vulnerable ghosts can be eaten while under the effects of a power pellet. It therefore defaults to evading the enemies whenever possible, which is illustrated in figure 3(a). It is not until those situations where the agent becomes trapped between two or more vulnerable ghosts that the policy shifts to actively eating ghosts. Being surrounded by ghosts is a somewhat rare occurrence, though, as the agent quickly learns to outpace them. We found that short-term increases in the learning rate helped to increase the adoption rate of this multi-modal behavior.

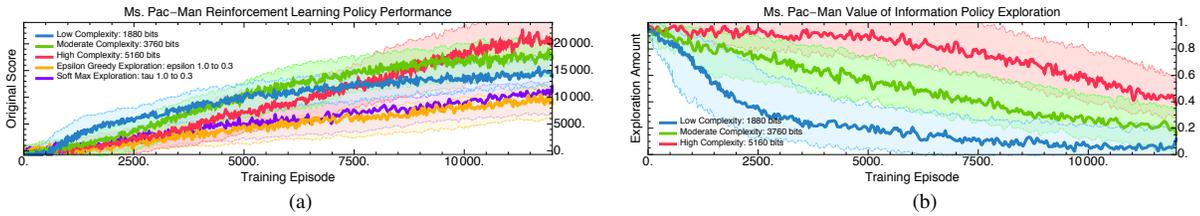

Figure 4: Plots of the value-of-information learning process. (a) Plot of the smoothed average original game score versus the number of episodes for different levels of complexity. This plot shows that lower complexity policies initially perform better than higher complexity ones due to a coarser quantization of the joint state-action space. The original scores versus the number of episodes for the best tested epsilon-greedy and soft-max exploration parameters are also shown in (a). (b) Plot of the smoothed, normalized average exploration amount versus the number of episodes for the value of information. This plot shows where the value of information is on the abnormal branch throughout training. Lower-risk policies move quickly away from the abnormal branch (one) toward the normal branch (zero). In both (a) and (b), results for the low-, moderate-, and high-complexity policies are represented using blue, green and red curves, respectively. In (a), results for epsilon-greedy and soft-max exploration and represented using yellow and purple curves, respectively. Darker lines show the average obtained over the independent trials. The lower and upper boundaries of the shaded area represent the 10 percentile ($z$ score: -1.25) and 90 percentile ($z$ score: 1.25) of the various quantities, respectively.

By around episode 7000 to 9500, the policies compel the agent to seek out ghosts whenever a power pellet is activated. This is demonstrated in figure 3(b), where the agent goes after two ghosts in rapid succession. We observed that only moderate- and high-complexity policies led the agents to chase vulnerable ghosts. High-complexity policies took the longest to implement this behavior, as shown in tables A.3(a)–A.3(c). They tended to lead to far superior strategies, though, after a sufficient number of episodes. This was due to a thorough investigation of the action space brought about by the high amount of risk. Low-complexity policies, in comparison, were too risk-adverse to stray from the initially learned behaviors. Moreover, the state-action space was too coarsely quantized to permit adequate action specialization to the relevant states.

These capabilities arose due to increased weighting of certain features. When comparing the feature weights in figures 1 and 2 to those in figure 3, it can be seen that the distance and direction to the nearest ghosts and power pellets become more prominent. The agent is not only trying to determine if it should head to the nearest power pellet, but also if there are any nearby ghosts for it to consume once it does. The feature associated with ghost vulnerability also had a larger influence over the action choices.

We observed that that moderate-complexity searches led to larger initial increases of these feature values than higher-complexity searches when this skill was implemented. Better scores are therefore produced during this time. This was to be expected, since the action space could be more quickly explored for the latter case. Moreover, once the agent found a state-action correspondence that led to good rewards, it would be more inclined to retain that correspondence versus choosing a different action. Repeatedly selecting different actions does offer some benefit, however. It leads to a more thorough investigation of the joint space and, with enough experience, a better understanding of the environment dynamics. High-complexity policies therefore will have higher feature weights than other types later in the learning process; they hence will yield superior multi-modal behaviors.

Removing the power-pellet-steps-remaining feature and restarting the simulations led to a modest learning-rate improvement for low-complexity policies. In fact, removing all of the power-pellet features allowed for the formation of pellet-seeking and ghost-avoidance behaviors more quickly in the low-risk case. For the higher-complexity cases, removing the former feature had a detrimental impact. The agent would chase after ghosts right until the effects of the power pellet wore off. They would therefore often be trapped by two or more nearby ghosts.

**Summary.** In this subsection, we covered basic game-playing behaviors that emerge during the simulations: seeking pellets and avoiding ghosts. We showed that this functionality arose due to an increased reliance on subsets of features used to represent the state space. The rate at which these features rose to prominence depended on the risk-taking parameter. Lower amounts of risk typically yielded faster feature-usage rates than higher amounts of risk. Only around 1000 episodes were needed before such behaviors could reliably clear the pellets on the initial levels. This was due to a coarse quantization of the action-state space, which could be thoroughly investigated in a relatively few number of episodes. Higher amounts of risk required more training episodes, typically on the order of 1000 to 2000. However, the behaviors could be refined to a greater degree for these cases, leading to better game-play performance.

We also covered two other behaviors could emerge: consuming fruit bonus items and chasing after vulnerable ghosts after eating power pellets. Both behaviors reliably materialized after around 3000 and 7000 episodes, respectively, for moderate and high policy complexities. This was due to a fine quantization of the action-state space, which permitted adequate specialization of action responses. It was also a byproduct of a high level of action-selection risk and hence experimenting with different strategies. Low complexity policies, in contrast, rarely implemented these two behaviors. This stemmed from an insufficient amount of action exploration.

#### 4.2.2 Policy Performance

The next aspect of the value of information that we consider is the relationship between risk and the obtainable score. We also analyze how such scores were obtained by looking at properties of the normal and abnormal branches, along with game-play metrics.

**Agent Score Analysis.** Figure 4(a) highlights the policies' original game scores across 12000 episodes for 25 independent trials. Figure 4(b) provides a plot of the normalized exploration amount on the abnormal branch across 12000 episodes for 25 independent trials. In figure 4(b), values close to one indicate a high amount of risk and hence exploration. Values close to zero correspond with low amounts of risk in the action-selection process and an exploitation of learned behaviors.

Figure 4(a) shows a sharp increase in rewards for all three policy complexities at the beginning of the learning process. This is due to the implementation of beneficial early-game behaviors. Figure 4(b) captures that these behaviors materialized due to a period of high exploration in the joint state-action space.

When considering an initial policy complexity of 5160 bits, the agent's performance is worse than for either 1880 or 3760 bits over the first 6000-7000 episodes. This is to be expected, as the policy solution space is being finely searched. The agent is therefore often attempting actions that deviate from the prescribed policy. Few behaviors are retained across multiple episodes at this stage. Figure 4(b) corroborates this claim. It shows that the high-complexity policy entries, and hence the action-value function estimates, are continuously revised. The search process is far into the abnormal branch during this time.

During the first 3000 episodes, high-complexity policies tend to leave more islands of pellets than the lower-complexity cases, as indicated in tables A.4(a)–A.4(c) of an associated online appendix. The agent also routinely out paces the ghosts, which is highlighted in table A.1(c). However, this is a marked decrease from the balanced-search case statistics given in table A.1(a) and A.1(b). Additionally, the agent also maneuvers to adjacent tunnels more slowly than the balanced-search case if it detects that one or more ghosts are present in the current tunnel. This is highlighted in tables A.5(a)–A.5(c). In short, high-complexity policies yield behaviors that lower the agent's chances of successfully completing the level early in the training process. This explains why the score is low for such policies at this stage of learning.

Given enough time to explore the space, though, high-complexity policies will perform just as well if not better than those of lower complexity. This typically occurs after 7000 to 8000 episodes. During this period, the high-complexity policies start to more aggressively chase after vulnerable ghosts over longer distances, as illustrated in tables A.3(a)–A.3(c). More ghosts are eaten per power pellet too, which can be seen from tables A.6(a)–A.6(c). The statistics in tables A.2(a)–A.2(c) indicate that the agents also pursue fruit bonus items over longer distances, thereby further improving the chance for large score improvements. Lastly, due to a refinement of the basic game-playing behaviors, the agents cleared more levels of pellets in a single episode. We refer to tables A.7(a)–A.7(c) for the details. All of these improvements were possible due to a thorough exploration of the action space. They also came about since the period of exploration was long enough to adjust to the multi-modal environment dynamics. Annealing the policy complexity additionally helped enable the agent to retain the highest-performing actions for each state group.

When considering an initial policy complexity 1880 bits, the performance improved in the first 3000 episodes compared to when 3760 or 5160 bits were used. The policy solution space is being coarsely searched in this case, which can be inferred from figure 4(b). The agent will only possess the most basic behaviors for winning the game, though, which include navigating to the nearest pellets and fleeing nearby ghosts. Rarely will these abilities be honed so that the agent consistently clears many levels as training progresses. This is because the amount of exploration is much lower in the later stages of learning, as indicated in figure 4(b), which prohibits the acquisition of new behaviors. The cost to transform the policy across episodes is also greater, which causes a shift away from the abnormal branch toward the normal branch.

During the first 3000 episodes, low-complexity policies outperform both moderate- and high-complexity policies. For instance, such policies typically remain more grid cells away from ghosts, which is noted in tables A.1(a)–A.1(c). It also takes, on average, fewer steps for the agent to evade ghosts by move to an empty tunnel. This is highlighted in tables A.5(a)–A.5(c). Both these statistics indicate that the agent will not be as easily caught in coordinated attacks from the ghosts. Lastly, such policies leave fewer islands of pellets than the moderate-complexity case, as shown in tables A.4(a)–A.7(c). Backtracking through the level is thus less necessary, which reduces the agent's risk of dying.

After about 3000 episodes, the performance of low-complexity policies lags behind the others. This is due to a few factors. First, moderate- and high-complexity policies tend to visit more levels per episode, as noted in tables A.7(a)–A.7(c). More pellets are consumed, which naturally improves the score. These higher-complexity policies also implement fruit-chasing behaviors. Since fruits are worth more in later levels, there is a potential for large score jumps. Secondly, low-complexity policies never capitalize on the relationship between power pellets and vulnerable ghosts. These policies therefore miss out on another score-increasing opportunity.

**Score Comparisons.** As shown in figure 4(a), the agent is able to accrue an average of about 20000 points after 12000 episodes. This score is competitive with the contest-winning scores of some of the earliest IEEE CEC and CIG Ms. Pac-Man competitions [3]. It is several thousand points away from the best scores obtained in later competitions, though; many of the competitor's approaches could reach scores in the 30000 point range.

It is important to note that such scores are not necessarily the best that can be obtained by our methodology. Figure 4(b) indicates that the highest-complexity policy is being adapted a fair amount once 12000 episodes is reached. It is possible that additional refinements to the game-play behaviors can be made after this point in the learning process.

We tested this hypothesis by increasing the learning period duration from 12000 to 25000 episodes for the highest-complexity policy. After 25000 episodes, the agent was achieving average scores of approximately 58000 points, indicating that it had adjusted better to the game-play dynamics. The best score that we obtained from this policy, out of ten independent games with three initial lives, was 69210. This score is competitive with other approaches that rely on a similar simulation environment. For exmaple, the best sores obtained in the IEEE CIG 2009, 2010, and 2011 competitions were 30010 (H. Matsumoto, et al.), 21250 (M. Emilio, et al.), and 26280 points (N. Ikehata and T. Ito), respectively. Ten Monte Carlo trials were run for each of the entries of these competitions, and only the highest score from the trials was used as a basis for comparison. Our scores also exceed those of the recent entry by Foderaro, et al. [32], who published an agent that could obtain 44630 points.

**Summary.** In this subsection, we assessed the scoring performance of the policies for different levels of risk. Our results illustrate that low-risk policies tended to increase the accrued rewards at a much faster pace than higher-risk policies during the first 1000 to 2500 episodes. This performance improvement rate was possible due to a coarse quantization of the action-state space, which could be explored quickly. We additionally showed that the score increases were connected with leaving fewer islands of pellets and quickly darting into tunnels that are not occupied by ghosts.

After this time, such low risk policies lagged, which was caused by a lack of adequate action specialization to different environment configurations. Both moderate- and high-risk policies dominated due the increased scores from consuming fruit bonus items and vulnerable ghosts. Eventually, high-risk policies produced the best results, as the agent's behaviors were steadily refined across episodes 7000 to 12000. Large score improvements were seen during this time. The improvement for moderate-risk policies, in contrast, slowed after around 10000 episodes. This occurred because such searches eventually became too risk adverse to yield significant behavior improvements.

These results signal that the risk hyperparameter directly influences many factors of the learning process. We will be using this knowledge in our future endeavors to determine how to automatically adjust the hyperparameter so as to optimally improve the score across each episode.

### 4.2.3 Methodological Comparisons

In the two previous subsections, we have detailed the effects of risk on the agent's performance. We now highlight how this performance compares with existing reinforcement-learning search mechanisms. We utilize two prominent stochastic schemes: epsilon-greedy exploration and soft-max-based exploration.

**Epsilon-Greedy Exploration Comparisons.** Epsilon-greedy exploration is based on the notion that random actions should be taken at random times. The choice of when to take a random action is governed by single parameter. It is well known that this type of approach can converge to optimal values, in certain situations, as the number of episodes grows [51]. For our application, this did not occur. The average scores are reported in table A.8(a) of an associated online appendix when annealing from an epsilon parameter value of 1.0 to 0.5. This corresponds to the situation of selecting a random action after every two steps. Increasing the final value of the annealed parameter to 0.7, which encourages a high degree of exploration, led to worse results. These results are presented in table A.8(b). Decreasing the final annealed parameter value to 0.3, which promotes exploitation, yielded a marginal improvement, as we show in table A.8(c); comparisons are provided in figure 4(a). Learning over anywhere from two to four times more episodes did little to enhance the agent's performance. Adjusting the learning rate and discount factor had only a marginal impact as well.

One of the changes that we found to work well involved a constant annealing of epsilon from 1.0 to 0.05 during training. This case corresponds to initially choosing a random action at every step to only choosing a random action every twenty steps. We found that this amount of exploration enabled the agent to implement basic game-play behaviors of pellet seeking and ghost avoidance at almost the same rate as policies using the value of information. This was because the policies would often retain good behaviors.

**Soft-Max Exploration Comparisons.** Part of the reason why epsilon-greedy exploration performs poorly is that there is no weighting of the actions being considered. The influence of the state-action pairs on the expected returns should, at the very least, guide the selection of appropriate actions, as in soft-max-based exploration. In our application, implementing soft-max-based exploration yielded modest benefits. When annealing the exploration-parameter value from 1.0 to 0.5, the average scores were higher than epsilon greedy across almost all of the episodes, as indicated in table A.9(a). This parameter value corresponds to an even mixture of exploration and exploitation. Biasing toward the best-performing action, with a final annealed parameter value of 0.3, led to slightly better scores. This is highlighted in table A.9(c); it is also presented in figure 4(a). Promoting near-equi-probable action selection, with a final annealed parameter value of 0.7, led to the rewards reported in table A.9(b). While these policies outperform those from epsilon-greedy exploration, they behaved worse than those produced by the value of information. Moreover, they did not implement any fruit-chasing capabilities.

Neither epsilon-greedy- nor soft-max-based policies implemented multi-modal agent behaviors well in the number of training episodes that we considered. The agents would often avoid vulnerable and invulnerable ghosts alike. This was largely expected, as such exploration schemes tend to define monolithic policies. Without either substantially more training episodes or the use of learning heuristics like action replay, it would be difficult for these schemes to promote the formation of multi-modal policies.

**Comparison Explanations.** For these simulations, value-of-information-based policies performed better because they induce a non-linear quantization of the state-action space according to the action-value function. The magnitude of the risk-taking parameter dictates the number of state clusters that are formed. The policy-complexity values that we considered were less than the state random-variable entropy, which led to a partitioning of the space. Much less effort was required to perform reinforcement learning over this partitioned space. Both epsilon-greedy and soft-max selection, in contrast, correspond to the value of information with a policy complexity equal to the state random-variable entropy. No quantization is performed in this case, and each state must be visited to be paired with an action. A great deal of agent-environment interactions are required for the agent to arrive at meaningful action-value function estimates.

Another reason why the value of information performed well is that it inherently accounts for the deviations in expected utility, or risk, when choosing actions. The value of information also weights action choices differently in risk-adverse situations compared to risk-favoring cases. Both of these properties ensure that the value of information is maximizing the action-value function over the potential outcomes at an optimal rate.

Many previous works suggest that risk is also connected with high-order culmulants of the utility function. In our case, we use a conditional Shannon entropy difference to model the risk, or uncertainty, associated with deviating from the supplied prior policy. It is well known that entropy accounts for all statistical moments [6]. Value-of-information-based agents hence operate with complete knowledge of the uncertainty associated with a particular action choice. Other exploration approaches often account for risk using, at most, only a single statistical moment. Some methods, such as epsilon-greedy, do not rely on any moments of the utility function. The rate at which such methods converge to optimal value functions can lag significantly behind the value of information, which we surmise is due to not accounting for uncertainty in the action-selection process.

**Summary.** We have shown that the value of information leads to the timely formation of high-performing policies. Comparatively, two of the most popular exploration heuristics, epsilon-greedy and soft-max selection, return sub-par policies. This is because they do not have the ability to quantize the state-action space and therefore must search over the entirety of it. It is also because these methods are not entirely risk sensitive. Although some methods, like soft-max selection, do weight action choices by their expected utility, they reason about risk in a symmetric manner. Asymmetry in risk-based decision making is needed for optimal decision-making under uncertainty [36].

In our future endeavors, we will demonstrate that this performance improvement occurs because such heuristics do not induce an aggregation of Markov chains underlying the Markov decision processes. The value of information, in contrast, performs this type of dynamics reduction, which further explains why it is quicker at uncovering good policies.

## 5  Conclusions

In this paper, we have considered an information-theoretic approach for either promoting or eschewing risky agent behaviors during reinforcement learning. Our approach is based on a value-of-information criterion, which optimally trades off between information and expected rewards.

The value of information can be utilized within a variety of dynamical-systems-based decision-making frameworks. When applied to Markov decision processes, the criterion quantifies the expected improvement in rewards that can be achieved for probabilistic policies of a specified complexity.

The value of information possesses a free hyperparameter whose value must be selected. This parameter dictates the amount of policy quantization. For the first form of this criterion, small parameter values correspond with low-complexity policies. It also corresponds with risk-avoiding agent behaviors. Higher parameter values lead to lower policy compression levels. Agents are more likely to frequently take risks in these cases, which can promote the formation of multi-modal strategies. What constitutes large and small values depends on the problem domain.

To gauge the performance of the value of information, we considered the creation of Ms. Pac-Man agents. We showed that the risk hyperparameter had a profound influence on the policy return quality for this game. High levels of risk yielded policies whose expected reward approached those for the optimal solution. They also facilitated the switching of game-play strategies, which, for this environment, involved chasing vulnerable ghosts. A great many learning episodes were needed to construct such multi-modal policies, though. Lower levels of risk allowed for the quick generation of policies that performed reasonably well. As the number of episodes increased, the policies' qualities lagged behind those found with higher levels of risk. They were also unable to reliably produce multi-modal game-play strategies.

These findings indicate that the value of information offers a good framework to understand the variables at play. We have only a single hyperparameter whose value must be chosen. This hyperparameter dictates a great deal of functionality, such as the amount of action-selection risk and hence the amount of exploration. Many other approaches for producing Ms. Pac-Man agents have multiple parameters and hyperparameters whose values must be chosen. Such a large number of parameters makes the connections between scores and variable values more difficult to understand. It can also impede the construction of automated, principled parameter-tuning methods.

In our simulations, we additionally compared the results of value-of-information-based exploration with two other reinforcement-learning search mechanisms: soft-max-based and epsilon-greedy action selection. Both schemes produced worse policies. They also did not implement multi-modal behaviors. This is because such selection procedures assume that the policies can be unboundedly complex. A great many episodes would therefore be needed to explore the joint state-action space. For many problems, value-of-information-based policies with finite complexity can approach the same levels of performance in fewer episodes. This is because not every state in a given problem may need to be paired with a unique action. It may hence be possible to group several states and assign them the same action without impacting performance.

These comparative results illustrate that the value-of-information criterion's merger of prospect theory and utility theory with uncertainty provides an excellent means of addressing the exploration dilemma. Neither epsilon-greedy exploration nor soft-max selection, in contrast, let alone many other search mechanisms, adequately account for uncertainty when choosing actions. They can therefore be ill equipped to understand the dynamics of highly complex, stochastic environments like the Ms. Pac-Man game world in only a few episodes.

Our emphasis in this paper has been on demonstrating that the value of information provides a principled trade-off between exploration and exploitation. Our future endeavors will involve extensions of this criterion to effectively address many classes of problems. One extension will a more principled selection of the criterion's free hyperparameter. Currently, we manually specified an initial value based upon the early simulation results. For certain domains, however, such trial-and-error-based testing may be prohibitively time consuming.

To sidestep this issue, we will investigate information-theoretic approaches for adapting the parameter value. A useful procedure may be to choose values that monotonically decrease the action-state entropy. By minimizing entropy, we are forcing the policies to eventually become deterministic. No further risks will need to be taken by the agent, as complete knowledge of that environment will have been obtained. Another possibility would be to utilize the policy cross-entropy to tune the hyperparameter value. Policy cross-entropy can be viewed as a bounded measure of how much the policy is changing across episodes. By adjusting the hyperparameter value so that the policy cross-entropy does not stagnate early in the learning process, we can be assured that a thorough exploration of the joint space will be being conducted. Moreover, high-performing policies should be found much more quickly than by a naive annealing procedures, such as the one that we considered.

IEEE TRANSACTIONS ON COMPUTATIONAL INTELLIGENCE AND ARTIFICIAL INTELLIGENCE (ACCEPTED) 18[11] M. Wittikamp, L. Barone, and P. Hingston, "Using NEAT for continuous adaptation and teamwork formation in Pac-Man," in *Proceedings of the IEEE Symposium on Computational Intelligence and Games (CIG)*, Perth, Australia, December 15-18 2008, pp. 234–242. Available: http://dx.doi.org/10.1109/CIG.2008.5035645

[12] M. Gallagher and M. Ledwich, "Evolving Pac-Man players: Can we learn from raw input?" in *Proceedings of the IEEE Symposium on Computational Intelligence and Games (CIG)*, Honolulu, HI, USA, April 1-5 2007, pp. 282–287. Available: http://dx.doi.org/10.1109/CIG.2007.368110

[13] R. Thawonmas and T. Ashida, "Evolution strategy for optimizing parameters in Ms. Pac-Man controller ICE Pambush 3," in *Proceedings of the IEEE Symposium on Computational Intelligence and Games (CIG)*, Copenhagen, Denmark, August 18-21 2010, pp. 235–240. Available: http://dx.doi.org/10.1109/ITW.2010.5593350

[14] A. M. Alhejali and S. M. Lucas, "Using a training camp with genetic programming to Ms. Pac-Man agents," in *Proceedings of the IEEE Symposium on Computational Intelligence and Games (CIG)*, Seoul, South Korea, August 31-September 3 2011, pp. 118–125. Available: http://dx.doi.org/10.1109/CIG.2011.6031997

[15] M. F. Brandstetter and S. Ahmadi, "Reactive control of Ms. Pac-Man using information retrieval based on genetic programming," in *Proceedings of the IEEE Symposium on Computational Intelligence and Games (CIG)*, Granada, Spain, September 11-14 2012, pp. 250–256. Available: http://dx.doi.org/10.1109/CIG.2012.6374163

[16] M. Gallagher and A. Ryan, "Learning to play Pac-Man: An evolutionary, rule-based approach," in *Proceedings of the IEEE Congress on Evolutionary Computation (CEC)*, Canberra, Australia, December 8-12 2003, pp. 2462–2469. Available: http://dx.doi.org/10.1109/CEC.2003.1299397

[17] P. Rohlfshagen and S. M. Lucas, "Ms. Pac-Man versus Ghost Team CEC 2011 competition," in *Proceedings of the IEEE Congress on Evolutionary Computation (CEC)*, New Orleans, LA, USA, June 5-8 2011, pp. 70–77. Available: http://dx.doi.org/10.1109/CEC.2011.5949599

[18] A. Fitzgerald and C. B. Congdon, "RAMP: A rule-based agent for Ms. Pac-Man," in *Proceedings of the IEEE Congress on Evolutionary Computation (CEC)*, Trondheim, Norway, May 18-21 2009, pp. 2646–2653. Available: http://dx.doi.org/10.1109/CEC.2009.4983274

[19] N. Wirth and M. Gallagher, "An influence map model for playing Ms. Pac-Man," in *Proceedings of the IEEE Symposium on Computational Intelligence and Games (CIG)*, Perth, Australia, December 15-18 2008, pp. 228–233. Available: http://dx.doi.org/10.1109/CIG.2008.5035644

[20] D. J. Gagne and C. B. Congdon, "FRIGHT: A flexible rule-based intelligent ghost team for Ms. Pac-Man," in *Proceedings of the IEEE Symposium on Computational Intelligence and Games (CIG)*, Granada, Spain, September 11-14 2012, pp. 273–280. Available: http://dx.doi.org/10.1109/CIG.2012.6374166

[21] P. Burrow and S. M. Lucas, "Evolution versus temporal difference learning for learning to play Ms. Pac-Man," in *Proceedings of the IEEE Symposium on Computational Intelligence and Games (CIG)*, Milan, Italy, September 7-10 2009, pp. 53–60. Available: http://dx.doi.org/10.1109/CIG.2009.5286495

[22] L. L. DeLooze and W. R. Viner, "Fuzzy $Q$-learning in a nondeterministic environment: Developing an intelligent Ms. Pac-Man agent," in *Proceedings of the IEEE Symposium on Computational Intelligence and Games (CIG)*, Milan, Italy, September 7-10 2009, pp. 162–169. Available: http://dx.doi.org/10.1109/CIG.2009.5286478

[23] S. Griffith, K. Subramanian, J. Scholz, C. Isbell, and A. L. Thomaz, "Policy shaping: Integrating human feedback with reinforcement learning," in *Advances in Neural Information Processing Systems (NIPS)*, C. J. C. Burges, L. Bottou, M. Welling, Z. Ghahramani, and K. Q. Weinberger, Eds. Cambridge, MA, USA: MIT Press, 2013, pp. 2625–2633.

[24] A. Vezhnevets, V. Mnih, S. Osindero, A. Graves, O. Vinyals, J. Agapiou, and K. Kavukcuoglu, "Strategic attentive writer for learning macro-actions," in *Advances in Neural Information Processing Systems (NIPS)*, D. D. Lee, M. Sugiyama, U. V. Luxburg, I. Guyon, and R. Garnett, Eds. Cambridge, MA, USA: MIT Press, 2016, pp. 3486–3494.

[25] J. Oh, X. Guo, H. Lee, R. L. Lewis, and S. Singh, "Action-conditional video prediction using deep networks in Atari games," in *Advances in Neural Information Processing Systems (NIPS)*, C. Cortes, N. D. Lawrence, D. D. Lee, M. Sugiyama, and R. Garnett, Eds. Cambridge, MA, USA: MIT Press, 2015, pp. 2863–2871.

[26] H. P. van Hasselt, A. Guez, M. Hessel, V. Mnih, and D. Silver, "Learning values across many orders of magnitude," in *Advances in Neural Information Processing Systems (NIPS)*, D. D. Lee, M. Sugiyama, U. V. Luxburg, I. Guyon, and R. Garnett, Eds. Cambridge, MA, USA: MIT Press, 2016, pp. 4287–4295.

[27] D. Robles and S. M. Lucas, "A simple tree search method for playing Ms. Pac-Man," in *Proceedings of the IEEE Symposium on Computational Intelligence and Games (CIG)*, Milan, Italy, September 7-10 2009, pp. 249–255. Available: http://dx.doi.org/10.1109/CIG.2009.5286469

[28] N. Ikehata and T. Ito, "Monte Carlo tree search in Ms. Pac-Man," in *Proceedings of the IEEE Symposium on Computational Intelligence and Games (CIG)*, Seoul, South Korea, August 31-September 3 2011, pp. 39–46. Available: http://dx.doi.org/10.1109/CIG.2011.6031987

[29] S. Samothrakis, D. Robles, and S. M. Lucas, "Fast approximate Max-n Monte Carlo tree search for Ms. Pac-Man," *IEEE Transactions on Computational Intelligence and AI in Games*, vol. 3, no. 2, pp. 142–154, 2011. Available: http://dx.doi.org/10.1109/TCIAIG.2011.2144597

[30] C. Browne, E. Powley, D. Whitehouse, S. Lucas, P. I. Cowling, P. Rohlfshagen, S. Tavener, D. Perez, S. Samothrakis, and S. Colton, "A survey of Monte Carlo tree search methods," *IEEE Transactions on Computational Intelligence and Artificial Intelligence in Games*, vol. 4, no. 1, pp. 1–49, 2012. Available: http://dx.doi.org/10.1109/TCIAIG.2012.2186810

[31] T. Pepels, M. H. M. Winands, and M. Lanctot, "Real-time Monte Caro tree search in Ms. Pac-Man," *IEEE Transactions on Computational Intelligence and AI in Games*, vol. 6, no. 3, pp. 245–257, 2014. Available: http://dx.doi.org/10.1109/TCIAIG.2013.2291577

[32] G. Foderaro, A. Swingler, and S. Ferrari, "A model-based cell decomposition approach to on-line pursuit-evasion path planning and the video game Ms. Pac-Man," in *Proceedings of the IEEE Symposium on Computational

**Results Supplement**

(a) Agent-Ghost Distance, Low-Complexity

| Episode | Avg. Distance |
| --- | --- |
| 0 to 1000 | $7.72 \pm 2.39$ cells |
| 1001 to 2000 | $7.91 \pm 2.54$ cells |
| 2001 to 3000 | $8.23 \pm 2.77$ cells |
| 3001 to 6000 | $9.75 \pm 2.91$ cells |
| 6001 to 9000 | $10.13 \pm 3.09$ cells |
| 9001 to 12000 | $10.28 \pm 3.18$ cells |

(b) Agent-Ghost Distance, Moderate-Complexity

| Episode | Avg. Distance |
| --- | --- |
| 0 to 1000 | $6.55 \pm 2.23$ cells |
| 1001 to 2000 | $7.96 \pm 2.56$ cells |
| 2001 to 3000 | $8.49 \pm 3.08$ cells |
| 3001 to 6000 | $9.15 \pm 3.57$ cells |
| 6001 to 9000 | $10.43 \pm 3.85$ cells |
| 9001 to 12000 | $11.18 \pm 4.20$ cells |

(c) Agent-Ghost Distance, High-Complexity

| Episode | Avg. Distance |
| --- | --- |
| 0 to 1000 | $5.96 \pm 2.62$ cells |
| 1001 to 2000 | $6.28 \pm 2.84$ cells |
| 2001 to 3000 | $8.83 \pm 2.92$ cells |
| 3001 to 6000 | $10.50 \pm 3.23$ cells |
| 6001 to 9000 | $11.32 \pm 4.38$ cells |
| 9001 to 12000 | $12.04 \pm 4.81$ cells |

Table A.1: Episode-averaged agent-ghost distance.

(a) Fruit-Chasing Distance, Low-Complexity

| Episode | Avg. Distance |
| --- | --- |
| 0 to 3000 | $0.00 \pm 0.00$ cells |
| 3001 to 6000 | $0.00 \pm 0.00$ cells |
| 6001 to 9000 | $0.54 \pm 0.27$ cells |
| 9001 to 12000 | $0.87 \pm 0.45$ cells |

(b) Fruit-Chasing Distance, Moderate-Complexity

| Episode | Avg. Distance |
| --- | --- |
| 0 to 3000 | $0.68 \pm 0.17$ cells |
| 3001 to 6000 | $2.75 \pm 1.92$ cells |
| 6001 to 9000 | $5.46 \pm 2.23$ cells |
| 9001 to 12000 | $8.09 \pm 2.84$ cells |

(c) Fruit-Chasing Distance, High-Complexity

| Episode | Avg. Distance |
| --- | --- |
| 0 to 3000 | $0.43 \pm 0.09$ cells |
| 3001 to 6000 | $1.54 \pm 0.97$ cells |
| 6001 to 9000 | $6.08 \pm 2.01$ cells |
| 9001 to 12000 | $10.61 \pm 3.78$ cells |

Table A.2: Episode-averaged distance that the agent will travel to seek out fruits.

(a) Vulnerable Ghost-Chasing Distance, Low-Complexity

| Episode | Avg. Distance |
| --- | --- |
| 0 to 3000 | $0.00 \pm 0.00$ cells |
| 3001 to 6000 | $0.00 \pm 0.00$ cells |
| 6001 to 9000 | $0.15 \pm 0.04$ cells |
| 9001 to 12000 | $0.24 \pm 0.09$ cells |

(b) Vulnerable Ghost-Chasing Distance, Moderate-Complexity

| Episode | Avg. Distance |
| --- | --- |
| 0 to 3000 | $0.35 \pm 0.08$ cells |
| 3001 to 6000 | $2.26 \pm 1.55$ cells |
| 6001 to 9000 | $6.81 \pm 2.13$ cells |
| 9001 to 12000 | $9.44 \pm 2.86$ cells |

(c) Vulnerable Ghost-Chasing Distance, High-Complexity

| Episode | Avg. Distance |
| --- | --- |
| 0 to 3000 | $0.27 \pm 0.11$ cells |
| 3001 to 6000 | $1.92 \pm 0.74$ cells |
| 6001 to 9000 | $6.19 \pm 2.15$ cells |
| 9001 to 12000 | $11.27 \pm 3.87$ cells |

Table A.3: Episode-averaged distance that the agent will travel to seek out the first vulnerable ghost after consuming a power pellet.

(a) Max Number of Pellet Islands, Low-Complexity

| Episode | Max Avg. Islands |
| --- | --- |
| 0 to 3000 | $6.54 \pm 2.65$ |
| 3001 to 6000 | $5.63 \pm 2.17$ |
| 6001 to 9000 | $5.14 \pm 2.06$ |
| 9001 to 12000 | $4.42 \pm 1.92$ |

(b) Max Number of Pellet Islands, Moderate-Complexity

| Episode | Max Avg. Islands |
| --- | --- |
| 0 to 3000 | $7.24 \pm 3.10$ |
| 3001 to 6000 | $6.13 \pm 2.85$ |
| 6001 to 9000 | $5.91 \pm 2.41$ |
| 9001 to 12000 | $4.05 \pm 1.73$ |

(c) Max Number of Pellet Islands, High-Complexity

| Episode | Max Avg. Islands |
| --- | --- |
| 0 to 3000 | $7.81 \pm 3.58$ |
| 3001 to 6000 | $7.29 \pm 2.96$ |
| 6001 to 9000 | $5.34 \pm 2.25$ |
| 9001 to 12000 | $3.79 \pm 1.16$ |

Table A.4: Episode-averaged maximum number of pellet islands that the agent will leave in an episode after losing a life. A pellet island is defined as an isolated group of pellets. Each level initially starts with a single island of pellets.

(a) Max Ghost-Avoidance Time, Low-Complexity

| Episode | Max Avg. Time |
| --- | --- |
| 0 to 3000 | $7.88 \pm 2.92$ steps |
| 3001 to 6000 | $6.41 \pm 2.70$ steps |
| 6001 to 9000 | $6.12 \pm 2.49$ steps |
| 9001 to 12000 | $5.93 \pm 2.27$ steps |

(b) Max Ghost-Avoidance Time, Moderate-Complexity

| Episode | Max Avg.Time |
| --- | --- |
| 0 to 3000 | $8.60 \pm 2.76$ steps |
| 3001 to 6000 | $7.27 \pm 2.53$ steps |
| 6001 to 9000 | $6.51 \pm 2.08$ steps |
| 9001 to 12000 | $5.18 \pm 1.81$ steps |

(c) Max Ghost-Avoidance Time, High-Complexity

| Episode | Max Avg. Time |
| --- | --- |
| 0 to 3000 | $8.91 \pm 2.97$ steps |
| 3001 to 6000 | $8.25 \pm 2.74$ cells |
| 6001 to 9000 | $7.32 \pm 2.22$ steps |
| 9001 to 12000 | $5.44 \pm 1.69$ steps |

Table A.5: Episode-averaged maximum number of actions needed for the agent to move to a tunnel that is not occupied by a ghost.

(a) Ghosts Eaten Per Power Pellet, Low-Complexity

| Episode | Avg. Num. Ghosts |
| --- | --- |
| 0 to 3000 | $0.00 \pm 0.00$ |
| 3001 to 6000 | $0.00 \pm 0.00$ |
| 6001 to 9000 | $0.00 \pm 0.00$ |
| 9001 to 12000 | $0.09 \pm 0.02$ |

(b) Ghosts Eaten Per Power Pellet, Moderate-Complexity

| Episode | Avg. Num. Ghosts |
| --- | --- |
| 0 to 3000 | $0.49 \pm 0.10$ |
| 3001 to 6000 | $0.87 \pm 0.16$ |
| 6001 to 9000 | $1.27 \pm 0.24$ |
| 9001 to 12000 | $1.42 \pm 0.39$ |

(c) Ghosts Eaten Per Power Pellet, High-Complexity

| Episode | Avg. Num. Ghosts |
| --- | --- |
| 0 to 3000 | $0.38 \pm 0.11$ |
| 3001 to 6000 | $1.09 \pm 0.20$ |
| 6001 to 9000 | $1.36 \pm 0.35$ |
| 9001 to 12000 | $1.64 \pm 0.48$ |

Table A.6: Episode-averaged number of ghosts eaten per power pellet.

(a) Number of Cleared Levels, Low-Complexity

| Episode | Avg. Num. Levels |
|---|---|
| 0 to 3000 | $2.17 \pm 0.51$ |
| 3001 to 6000 | $3.46 \pm 0.91$ |
| 6001 to 9000 | $4.85 \pm 1.24$ |
| 9001 to 12000 | $5.64 \pm 1.18$ |

(b) Number of Cleared Levels, Moderate-Complexity

| Episode | Avg. Num. Levels |
|---|---|
| 0 to 3000 | $1.74 \pm 0.62$ |
| 3001 to 6000 | $4.58 \pm 1.11$ |
| 6001 to 9000 | $5.36 \pm 1.44$ |
| 9001 to 12000 | $6.41 \pm 1.75$ |

(c) Number of Cleared Levels, High-Complexity

| Episode | Avg. Num. Levels |
|---|---|
| 0 to 3000 | $1.32 \pm 0.33$ |
| 3001 to 6000 | $3.04 \pm 1.20$ |
| 6001 to 9000 | $4.72 \pm 1.76$ |
| 9001 to 12000 | $6.87 \pm 1.94$ |

Table A.7: Episode-averaged number of levels cleared before all lives run out.

(a) Epsilon-Greedy Exploration Performance, $\epsilon_0 = 1.0$ to $\epsilon_{12000} = 0.5$

| Episode | Avg. Score |
|---|---|
| 0 to 3000 | $2134.10 \pm 950.24$ |
| 3001 to 6000 | $3391.53 \pm 1234.09$ |
| 6001 to 9000 | $4832.15 \pm 1599.87$ |
| 9001 to 12000 | $6061.58 \pm 1821.46$ |

(b) Epsilon-Greedy Exploration Performance, $\epsilon_0 = 1.0$ to $\epsilon_{12000} = 0.7$

| Episode | Avg. Score |
|---|---|
| 0 to 3000 | $2078.06 \pm 823.81$ |
| 3001 to 6000 | $2400.48 \pm 1162.98$ |
| 6001 to 9000 | $3527.51 \pm 1426.52$ |
| 9001 to 12000 | $5417.72 \pm 1313.65$ |

(c) Epsilon-Greedy Exploration Performance, $\epsilon_0 = 1.0$ to $\epsilon_{12000} = 0.3$

| Episode | Avg. Score |
|---|---|
| 0 to 3000 | $2274.92 \pm 861.44$ |
| 3001 to 6000 | $4689.47 \pm 1595.15$ |
| 6001 to 9000 | $6845.03 \pm 2023.76$ |
| 9001 to 12000 | $8327.89 \pm 2818.82$ |

Table A.8: Episode-averaged score for epsilon-greedy exploration

(a) Soft-Max Exploration Performance, $\tau_0^{-1} = 1.0$ to $\tau_{12000}^{-1} = 0.5$

| Episode | Avg. Score |
|---|---|
| 0 to 3000 | $2299.65 \pm 835.89$ |
| 3001 to 6000 | $4310.20 \pm 1108.55$ |
| 6001 to 9000 | $6548.37 \pm 1583.00$ |
| 9001 to 12000 | $8743.26 \pm 1932.13$ |

(b) Soft-Max Exploration Performance, $\tau_0^{-1} = 1.0$ to $\tau_{12000}^{-1} = 0.7$

| Episode | Avg. Score |
|---|---|
| 0 to 3000 | $2346.87 \pm 985.18$ |
| 3001 to 6000 | $4340.65 \pm 1344.04$ |
| 6001 to 9000 | $5952.14 \pm 1731.70$ |
| 9001 to 12000 | $6238.92 \pm 1986.54$ |

(c) Soft-Max Exploration Performance, $\tau_0^{-1} = 1.0$ to $\tau_{12000}^{-1} = 0.3$

| Episode | Avg. Score |
|---|---|
| 0 to 3000 | $2573.43 \pm 821.67$ |
| 3001 to 6000 | $4985.13 \pm 1023.71$ |
| 6001 to 9000 | $7232.78 \pm 1585.32$ |
| 9001 to 12000 | $10292.59 \pm 2237.99$ |

Table A.9: Episode-averaged score for soft-max exploration